# Personalized Cinemagraphs using Semantic Understanding and Collaborative Learning*

Tae-Hyun Oh[1,2][†]   Kyungdon Joo[2][†]   Neel Joshi[3]   Baoyuan Wang[3]   In So Kweon[2]   Sing Bing Kang[3]
[1]MIT CSAIL, Boston, MA   [2]KAIST, South Korea   [3]Microsoft Research, Redmond, WA

## Abstract

*Cinemagraphs are a compelling way to convey dynamic aspects of a scene. In these media, dynamic and still elements are juxtaposed to create an artistic and narrative experience. Creating a high-quality, aesthetically pleasing cinemagraph requires isolating objects in a semantically meaningful way and then selecting good start times and looping periods for those objects to minimize visual artifacts (such a tearing). To achieve this, we present a new technique that uses object recognition and semantic segmentation as part of an optimization method to automatically create cinemagraphs from videos that are both visually appealing and semantically meaningful. Given a scene with multiple objects, there are many cinemagraphs one could create. Our method evaluates these multiple candidates and presents the best one, as determined by a model trained to predict human preferences in a collaborative way. We demonstrate the effectiveness of our approach with multiple results and a user study.*

## 1. Introduction

With modern cameras, it is quite easy to take short, high resolution videos or image bursts to capture the important and interesting moments. These small, dynamic snippets of time convey more richness than a still photo, without being as heavyweight as a longer video clip. The popularity of this type of media has spawned numerous approaches to capture and create them. The most straightforward methods make it as easy to capture this imagery as it is to take a photo (e.g., Apple Live Photo). To make these bursts more compelling and watchable, several techniques exist to stabilize (a survey can be found in [33]), or loop the video to create video textures [29] or "cinemagraphs" [1], a media where dynamic and still elements are juxtaposed, as a way to focus the viewer's attention or create an artistic effect.

The existing work in the space of cinemagraph and live image capture and creation has focused on ways to ease user burden, but these methods still require significant user control [2, 17]. There are also methods that automate the creation of the loops such that they are the most visually seamless [23], but they need user input to create aesthetic effects such as cinemagraphs.

We propose a novel, scalable approach for automatically creating semantically meaningful and pleasing cinemagraphs. Our approach has two components: (1) a new computational model that creates meaningful and consistent cinemagraphs using high-level semantics and (2) a new model for predicting person-dependent interestingness and visual appeal of a cinemagraph given its semantics. These two problems must be considered together in order to deliver a practical end-to-end system.

For the first component, our system makes use of semantic information by using object detection and semantic segmentation to improve the visual quality of cinemagraphs. Specifically, we reduce artifacts such as whole objects being separated into multiple looping regions, which can lead to tearing artifacts.

In the second component, our approach uses semantic information to generate a range of candidate cinemagraphs, each of which involves animation of a different object, *e.g.*, tree or person, and uses a machine learning approach to pick which would be most pleasing to a user, which allows us to present the most aesthetically pleasing and interesting cinemagraphs automatically. This is done by learning a how to rate a cinemagraph based on interestingness and visual appeal. Our rating function is trained using data from an extensive user study where subjects rate different cinemagraphs. As the user ratings are highly subjective, due to individual personal preference, we propose a collaborative filtering approach that allows us to generalize preferences of sub-populations to novel users. The overall pipeline of our system is shown in Fig. 1.

In summary, our technical contributions include: (1) a novel algorithm for creating semantically meaningful cinemagraphs, (2) a computational model that learns to rate (i.e., predict human preference for) cinemagraphs, and (3) a collaborative filtering approach that allows us to generalize and predict ratings for multiple novel user populations.

*Acknowledgment   We would like to thank all the participants in our user study. We are also grateful to Jian Sun and Jinwoo Shin for the helpful discussions. This work was mostly done while the first author was an intern at Microsoft Research, Redmond. It was completed at KAIST with the support of the Technology Innovation Program (No. 10048320), which is funded by the Korean government (MOTIE).

[†]The first and second authors contributed equally to this work.

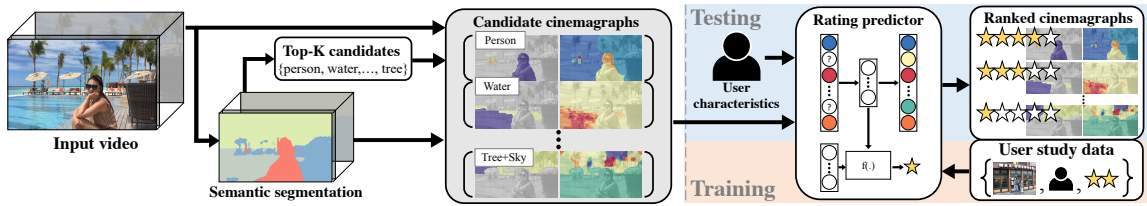

Figure 1: Overview of our semantic aware cinemagraph creation and suggestion system: 1) applying a semantic segmentation on the input video to recover semantic information, 2) selecting top-$K$ candidate objects, each of which will be dynamic in a corresponding candidate cinemagraph, 3) solving semantic aware Markov Random Field (MRF) for multiple candidate cinemagraph generation (Sec. 3). 4) selecting or ranking the best candidate cinemagraphs by a model learned to predict subjective preference from a database we acquire of user preferences for numerous cinemagraphs in an off-line process (Sec. 4).

## 2. Related Work

There is a range of types of imagery that can be considered a "live image", "live photo", or "living portrait". In this section, we briefly survey techniques for creating these types imagery, categorized roughly as video textures (whole frame looping), video looping (independent region looping), and content-based animation (or cinemagraphs).

**Video Textures** Video textures [29, 20, 24, 10] refer to the technique of optimizing full-frame looping given a short video. It involves the construction of a frame transition graph that minimizes appearance changes between adjacent frames. While the above methods are restricted to frame-by-frame transition of a video, the notion of video re-framing has inspired many video effect applications, *e.g.*, independent region-based video looping and cinemagraphs.

**Video Looping** Liao *et al*. [23] developed an automatic video-loop generation method that allows independently looping regions with separate periodicity and starting frames (optimized in a follow-up work [22]). The representation used in [22, 23] conveys a wide spectrum of dynamism that a user can optionally select in the generated video loop. However, the output video loop is generated without any knowledge of the scene semantics; the dynamics of looping is computed based on continuity in appearance over space and time. This may result in physically incoherent motion for a single object region (e.g., parts of a face may be animated independently). Our work builds directly on these approaches, by incorporating semantic information into cost functions.

**Interactive Cinemagraph Creation** The term "cinemagraph" was coined and popularized by photographer Jamie Beck and designer Kevin Burg [1], who used significant planning and still images shot with a stationary camera for creating cinemagraphs.

A number of interactive tools have been developed to make it easier to create cinemagraphs [35, 17, 2]. These approaches focus on developing a convenient interactive representation to allow user to composite a cinemagraph by manual strokes. Commercial and mobile apps such as Microsoft Pix, Loopwall, Vimeo's Echograph[1] and Flixel's Cinemagraph Pro[2] are also available, with varying degrees of automation. The primary difference between all these previous works and ours is that user input is not necessary for our method to create a cinemagraph effect.

**Automatic and Content-based Creation** Closely related to our work are techniques that perform automatic cinemagraph creation in a restricted fashion [40, 7, 39, 3, 30].

Bai *et al*. [3] track faces to create portrait cinemagraphs, while Yeh *et al*. [40, 39] characterize "interestingness" of candidate regions using low-level features such as cumulative motion magnitudes and color distinctness over sub-regions. More recently, Sevilla-Lara *et al*. [30] use non-rigid morphing to create a video-loop for the case of videos having a contiguous foreground that can be segmented from its background. Yan *et al*. [38] create a cinemagraph from a video (captured with a moving camera) by warping to a reference viewpoint and detecting looping regions as those with static geometry and dynamic appearance.

By comparison, our method is not restricted to specific target objects; we generate a cinemagraph as part of an optimization instead of directly from low-level features or very specific objects (*e.g.*, faces [3]). Our approach is to produce independent dynamic segments as with Liao *et al*. [23], but we encourage them to correspond as much as possible with semantically clustered segments. Given the possible candidates, each with a different looping object, we select the best cinemagraph by learned user preferences.

**Rating of Videos and Cinemagraphs** There are a few approaches to rank or rate automatically-generated videos. Gygli *et al*. [15] propose an automatic GIF generation method from a video, where it suggests ranked segments from a video in an order of popularity learned from GIFs on the web; however, their method does not actually *generate* an animated GIF or a video loop. Li *et al*. [21] create a benchmark dataset and propose a method to rank animated GIFs, but do not create them. Chan *et al*. [7] rank scene "beauty" in cinemagraphs based on low-level information (the size of the region of interest, motion magnitude, and duration of motion). We are not aware of any work that rates cinemagraphs based on user and high-level visual contexts.

---
[1] https://vimeo.com/echograph
[2] http://www.flixel.com/

## 3. Semantic Aware Cinemagraph Generation

A semantic segmentation of the scene allows us to model semantically meaningful looping motion in cinemagraph. In the following sections, we describe how we extract the semantic information for cinemagraph, and then how we instill it into an MRF optimization.

Note that throughout this paper, we assume that the input video is either shot on a tripod or stabilized using off-the-shelf video stabilization (*e.g.*, Adobe After Effect). Due to the space limit, we present details, *e.g.*, implementation, all the parameter values we used and setups if not specified, in the supplementary material.

### 3.1. Semantic Information Extraction

**Semantic Segmentation** We use semantic segmentation responses obtained from a model, FCN-8 [25][3] learned with PASCAL$_{\text{Context}}$[27], which predicts 60-classes per pixel. We run it on each frame of the input video independently, which forms the semantic response $F \in [0,1]^{C \times S \times T}$, where $C$, $S$ and $T$ denote the numbers of channels (or semantic category), spatial pixels and input video frames, respectively.

Empirically, we found that using macro-partitioned semantic categories causes over-segmentation, which is often undesirable for cinemagraph generation. We re-define categories that exhibit different types of cinemagraph motions and alleviate FCN's imperfect prediction that are easily confused by FCN. We combined some categories to generate a smaller number of representative higher-level categories which are roughly classified by similar semantics as well as similar cinemagraph motion characteristics, *e.g.*, {ground, floor, sidewalk} to be in *background*. We reduced the number of categories from 60 to 32 including background ($C$=32); all these mapping of categories are listed in the supplementary.

**Top-$K$ Candidate Label Selection** Unfortunately, this 32-dimensional (in short, dim.) feature introduces significant computational complexity in subsequent optimization. To reduce the complexity and memory usage, we only store the top-$K$ class responses to form semantic response. These top-$K$ classes are used in the optimization described later and determining what objects should be dynamic (*i.e.*, looping) in each candidate cinemagraph.

We select the top-$K$ by the number of pixels associated with each category with simple filtering. The procedure to select candidate objects is as follows:

1. Given $F \in [0,1]^{C \times S \times T}$, construct a global histogram $h_g(c) = \sum_{x,t} \delta [c = \mathrm{argmax}_{c'} F(c', x, t)]$, where $\delta[\cdot]$ denotes the indicator function to return 1 for true argument, otherwise 0,
2. Discard classes from $h_g$ that satisfy the following criteria:
   (a) Static object categories with common sense (*i.e.*, objects that do not ordinarily move by themselves, such as roads and buildings. The full lists are in the supplementary),
   (b) Object classes of which the standard deviation of intensity variation across time is ≤0.05 (*i.e.*, low dynamicity),
   (c) Object classes of which connected component blob sizes are too small on average (≤20×20 pixels),
3. Pick top-$K$ labels which are $K$ highest values in the histogram $h_g$, and with this, pick the channel dim. of $F$ to be $K$ as $F \in [0,1]^{K \times S \times T}$. We set $K = 4$.[4]

**Spatial Candidate Map $\pi$** Given top-$K$ candidate objects, we maintain another form of candidate information that allows our technique to decide which regions should appear as being dynamic in each candidate cinemagraph.

We use a rough per-pixel binary map $\pi_i \in \{0,1\}^S$ for each category $i$. Let $m[\cdot]:\{1,\cdots,K\} \to \{1,\cdots,C\}$ be the mapping from an index of the top-$K$ classes to an original class index. Then, we compute $\pi_{m[k]}$ by thresholding the number of occurrences of the specified candidate object $k$ across time as $\pi_{m[k]}(x) = \delta[h_t(k,x) \geq thr.]$, where $h_t(k,x) = \sum_t \delta[k = \mathrm{argmax}_{k'} F(k',x,t)]$ is a histogram across the temporal axis. The candidate region information from $\pi$ is propagated through subsequent MRF optimization.

### 3.2. Markov Random Field Model

Our MRF model builds on Liao *et al.* [23]. Their approach solves for an optimal looping period $p_x$ and start frame $s_x$ at each pixel, so that the input RGB video $\vec{V}(x,t) \in [0,1]^3$ is converted to an endless video-loop $\vec{L}(x,t) = \vec{V}(x,\phi(x,t))$ with a time-mapping function $\phi(x,t) = s_x + (t - s_x) \bmod p_x$. Following this formulation, we formulate the problem as 2D MRF optimization.

Liao *et al.*'s approach uses terms to minimize color difference between immediate spatiotemporal pixel neighbors, but it does not incorporate any high level information. Thus, while the resulting loops are seamless in terms of having minimal color differences of neighbors, it is common that the resulting video loops have significant artifacts due to the violation of semantic relationships in the video, *e.g.*, parts of objects like animal or person are often broken apart in resulting video-loops, which looks unnatural and awkward.

We extend the method of Liao *et al.* such that semantic consistency is also considered in the energy terms of the optimization along with photometric consistency. In addition to creating results that have fewer semantic-related artifacts, we use the semantically meaningful segments to create a variety of cinemagraph outputs where we can control the dynamic/static behavior on a per-object basis. Lastly, we adaptively adjust parameters according to semantic contexts, *e.g.*, enforce greater spatial consistency for a person, and require less consistency for stochastic non-object textures such as water and grass.

---
[3] We explain with FCN as a reference in this work, but it can be seamlessly replaced with an alternative one and all the technical details remain same.

[4] It has practical reasons: (1) A multiple of 4 allows word alignment for the memory bus. Bus transfer speed is important because the semantic feature vector is frequently evaluated during optimization. (2) Through many experiments, we found that four categories are enough to cover a wide range of dynamic scenes.

**Cost Function** Denoting start frames $\mathbf{s} = \{s_x\}$, periods $\mathbf{p} = \{p_x\}$, and labels $\mathbf{l} = \{l_x\}$, where $l_x = \{p_x, s_x\}$, we formulate the semantic aware video-loop problem as:

$$\operatorname*{argmin}_{\mathbf{s},\mathbf{p}} \sum_x \{E_{\text{temp.}}(l_x) + \alpha_1 E_{\text{label}}(l_x) + \alpha_2 \sum_{z \in \mathcal{N}(x)} E_{\text{spa.}}(l_x, l_z)\}, \quad (1)$$

where $z \in \mathcal{N}(x)$ indicates neighbor pixels. The basic ideas for the label term $E_{\text{label}}$, spatial and temporal consistency terms $E_{\text{spa.}}$ and $E_{\text{temp.}}$ are the same with those described in [23]. However, there are significant differences in our work, *i.e.*, our semantic aware cost function.

**Hyper-Classes for Semantic Aware Cost** Our empirical observation is that depending on types of object motion characteristics, qualities of resulting cinemagraphs vary as mentioned above. In this regard, a single constant value for each parameter in cost function limits the extent of its applicability. To allow the object specific adaptation, we control the dynamicity of resulting loops according to the class.

Assigning object dependent parameters for all the classes leads to parameter tuning on the high dimension parameter space, which is challenging. As a trade-off, we use another set of hyper-class by simply classifying $C$-classes into *natural / non-natural* texture to encourage the diversity of loop labels or to synchronize loop labels, respectively. The natural set $\mathcal{H}_{\text{nat.}}$ denotes the objects like tree, water, grass, waterfall, *etc.*, which are natural objects that have textual motion easily loopable and generally require less spatial coherence. The non-natural set $\mathcal{H}_{\text{non.}}$ denotes the objects like a person, animal, car, *etc.*, which have rigid or non-rigid motion and are very sensitive to incoherence. The full natural and non-natural category list is in the supplementary. The separation into "natural" and "non-natural" empirically allows us to enjoy few parameters but enough adaptation effectively.

**Temporal consistency term** Both consistency terms incorporate semantic and photometric consistency measures. The term $E_{\text{temp.}}$ measures the consistency across the loop start frame $s_x$ and the end frame $s_x + p_x$ as

$$E_{\text{temp.}}(l_x) = \gamma_t(x) \left[(1-w)\Phi_V(x) + w\Phi_F(x)\right], \quad (2)$$

where $w$ is the semantic balance parameter, the temporal photometric consistency $\Phi_V(x)$ and the temporal semantic consistency $\Phi_F(x)$ are defined as follows:

$$\Phi_V(x) = \tfrac{1}{3}\left(\begin{array}{c}\|\vec{V}(x,s_x) - \vec{V}(x,s_x+p_x)\|^2 + \\ \|\vec{V}(x,s_x-1) - \vec{V}(x,s_x+p_x-1)\|^2\end{array}\right),$$

$$\Phi_F(x) = \tfrac{1}{K}\left(\begin{array}{c}\|\vec{F}(x,s_x) - \vec{F}(x,s_x+p_x)\|^2 + \\ \|\vec{F}(x,s_x-1) - \vec{F}(x,s_x+p_x-1)\|^2\end{array}\right),$$

so that the loop is not only visually loopable, but also semantically loopable. We represent the semantic response $F$ in a vector form, $\vec{F}(x,t) \in [0,1]^K$.[5] The factor $\gamma_t(x)$ [20, 23] is defined as

$$\gamma_t(x) = 1/\left(1 + \lambda_t(x) \operatorname{MAD}_{t'}\|\vec{V}(x,t') - \vec{V}(x,t'+1)\|\right). \quad (3)$$

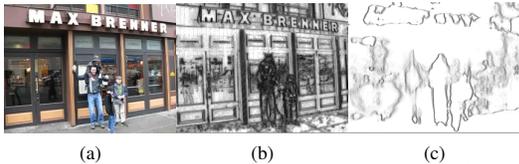

(a) (b) (c)

Figure 2: Comparison on connectivity potential $\gamma_s(x,z)$. (a) Selected frame. (b) $\gamma_s(x,z)$ in Liao *et al.* [23] (deviation of intensity difference across time). (b) Our version of $\gamma_s(x,z)$ (difference of semantic label occurrence distribution).

This factor estimates temporal intensity variation at $x$ based on the median absolute deviation (MAD). The factor $\gamma_t(x)$ slightly relaxes $E_{\text{temp.}}$ when the intensity variation is large, based on the observation that looping discontinuities are less perceptible in that case. The spatially varying weight $\lambda_t(x)$ is determined depending on semantic information as $\lambda_t(x) = 125$ if $(\vee_{i \in \mathcal{H}_{\text{nat.}}} \pi_i(x)) = 1$, where $\vee$ denotes the logical disjunction operator, otherwise it is half the value. By this, we reduce $\gamma_t(x)$ for the natural objects, as the loop discontinuity is less perceptible for the natural one.

**Spatial consistency term** The term $E_{\text{spa.}}$ also measures semantic and photometric consistency between neighbors as well. Specifically, $E_{\text{spa.}}$ is defined as

$$E_{\text{spa.}}(l_x, l_z) = \gamma_s(x,z)\left[(1-w)\Psi_V(x,z) + w\Psi_F(x,z)\right]. \quad (4)$$

The spatial photometric consistency $\Psi_V(x,z)$ and the spatial semantic consistency $\Psi_F(x,z)$ are defined as follows:

$$\Psi_V(x,z) = \tfrac{1}{3 \cdot \text{LCM}} \sum_{t=0}^{T-1}\left(\begin{array}{c}\|\vec{V}(x,\phi(x,t)) - \vec{V}(x,\phi(z,t))\|^2 + \\ \|\vec{V}(z,\phi(x,t)) - \vec{V}(z,\phi(z,t))\|^2\end{array}\right),$$

$$\Psi_F(x,z) = \tfrac{1}{K \cdot \text{LCM}} \sum_{t=0}^{T-1}\left(\begin{array}{c}\|\vec{F}(x,\phi(x,t)) - \vec{F}(x,\phi(z,t))\|^2 + \\ \|\vec{F}(z,\phi(x,t)) - \vec{F}(z,\phi(z,t))\|^2\end{array}\right),$$

where LCM is the least common multiple of per-pixel periods [23]. This cost can be evaluated efficiently by separating cases *w.r.t.* $l_x$ and $l_z$ and using an integral image technique in a constant time similar to Liao *et al.* [23].

We also define the connectivity potential, $\gamma_s(x,z)$, in a semantic aware way, to maintain coherence within objects. We introduce a label occurrence $\vec{h}_t(x) = [h_t(k,x)]_{k=1}^K$, where the histogram $h_t(k,x)$ was defined in Sec. 3.1. If two histograms between neighbor pixels are similar, it indicates that two pixels have a similar semantic occurrence behavior. We measure the connectivity potential by computing the difference of semantic label occurrence distribution as

$$\gamma_s(x, z) = 1 \Big/ \Big(1 + \lambda_s \|\hat{h}_t(x) - \hat{h}_t(z)\|_2\Big), \quad (5)$$

where $\hat{h}_t(\cdot)$ is the normalized version of $\vec{h}_t(x)$. As shown in Fig. 2, it preserves real motion boundaries better than the one proposed by Liao *et al.*

---
[5] When feeding semantic response $F$ into the subsequent optimization, we re-normalize each vector across the channel axis to sum to one.

**Algorithm 1** Procedure for Candidate Cinemagraph Generation.
1: **Input :** Video, semantic responses, spatial candidate map $\pi$.
2: Stage 1 (Initialization): Solve MRFs for **s**, given each $p>1$ fixed (*i.e.*, $\mathbf{s}^*_{|p}$).
3: (Multiple Candidate Cinemagraph Generation)
4: **for** each candidate label $\mathbb{ID}$ **do**
5:     Stage 2: Solve MRF for $\{\mathbf{p}>1, \mathbf{s}'\}$ given $\mathbb{ID}$, where each $p_x$ is paired as $(p_x, s^*_{x|p_x})$ from the step 2, $s'_x$ denotes all possible frames for the static case, $p=1$.
6:     Stage 3: Solve MRF for **s** given $\mathbb{ID}$ and fixed $\{\mathbf{p}^*\}$.
7:     Render the candidate cinemagraph result as described in Liao *et al.* [23, 22].
8: **end for**
9: **Output :** Candidate cinemagraphs.

**Label term** We define the label term, $E_\text{label}$, to assign an object-dependent spatial penalty in addition to discouraging a trivial all-static solution as in Liao *et al.* This is key in generating object-specific candidate cinemagraphs that allows us to vary which objects are static vs. looping.

Our label term $E_\text{label}$ is defined as:

$$E_\text{label}(l_x) = \begin{cases} E_\text{static}(x) \cdot \delta[\pi_{\mathbb{ID}}(x)], & p_x=1, \\ \alpha_\infty \cdot \delta[\vee_{i \in \mathcal{H}_\text{nat.}} \pi_i(x)], & 1<p_x \leq P_\text{short}, \\ 0, & P_\text{short} < p_x, \end{cases} \quad (6)$$

where $\mathbb{ID}$ represents the current target candidate category index the algorithm will generate, and $P_\text{short}$ defines the range of short periods. The label term $E_\text{label}$ has three cases. When $p_x = 1$, *i.e.*, static, the cost imposes the static penalty $E_\text{static}$ only when the semantic index at the pixel is the target label we want to make it dynamic. The static term $E_\text{static}(x)$ is defined as

$$E_\text{static}(x) = \alpha_\text{sta.} \min(1, \lambda_\text{sta.} \text{MAD}_{t'} \|N(x,t') - N(x, t'+1)\|), (7)$$

where $N$ is a Gaussian-weighted spatio-temporal neighborhood. The static term $E_\text{static}$ penalizes significant temporal variance of the pixels neighborhood in the input video, and also prevents a trivial solution which assigns all the pixel to be static that attains perfect spatio-temporal consistency.

We also observe that long periods look more natural for natural objects. To encourage long period, we add high penalty on natural object regions for short period labels ($1 < p_x \leq P_\text{short}$) with a large $\alpha_\infty$. Otherwise, $E_\text{label}$ is 0.

### 3.3. Optimization Procedure

The multi-label 2D MRF optimization in Eq. (1) can be solved by $\alpha$-expansion graph cut [19]. Due to the size of the label space, *i.e.*, $|\mathbf{s}| \times |\mathbf{p}|$, directly optimizing Eq. (1) may stuck in poor local minima. This is because a graph cut $\alpha$-expansion only deals with a single new candidate label at a time. Also, a video clip typically consists of mul-tiple semantic regions, whereby several candidate cinemagraphs are generated. We present an efficient procedure for multiple candidates in Alg. 1, which is regarded as a block coordinate descent. The stages (1) and (2) in Alg. 1 are similar to the procedure of Liao *et al.* [23] except the $\mathbb{ID}$ dependency involved. Moreover, due to the restriction of the paired label, $s_{x|p}$, in the stage (1), the solution can be still restricted up to the stage (2); hence we additionally introduce the stage (3).

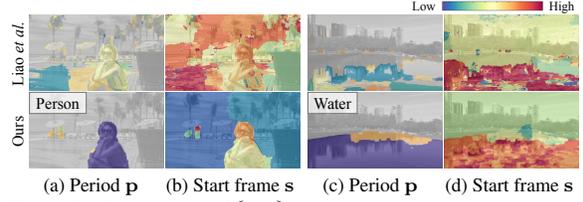

(a) Period **p**   (b) Start frame **s**   (c) Period **p**   (d) Start frame **s**

Figure 3: Visualization of $\{\mathbf{p},\mathbf{s}\}$ estimated by Liao *et al.* [23] (top) and ours (bottom). Values of **p** and **s** are presented by a color map on the top right corner, with gray indicating static pixels.

Since the terms related to candidate-specific regularization by $\mathbb{ID}$ are not involved in the stage (1), the initial paired label sets $\{(p_x, s_{x|p_x})\}$ obtained from the stage (1) are shared across all other stages. The complexities of each stage are proportion to the number of labels: $|\mathbf{s}|$, $|\mathbf{p}| + |\mathbf{s}|$ and $|\mathbf{s}|$ in the stages (1,2) and (3) respectively, which are significantly lower than directly optimizing the problem with $|\mathbf{p}| \times |\mathbf{s}|$ labels. The number of total candidate cinemagraphs generated is restricted to $K$. To obtain more diverse candidates, we allow the target $\mathbb{ID}$ to involve combination of multiple objects, *e.g.*, {Person, Tree} in $\mathbb{ID}$, so that both are dynamic in a candidate cinemagraph.

Fig. 3 visualizes the labels $\{\mathbf{p},\mathbf{s}\}$ obtained by our semantic-based cinemagraphs, which show strong spatial coherence along the semantic regions.

## 4. Learning to Predict Human Preference

Given a set of candidate cinemagraphs generated from a video clip, we want to automate suggesting a single *best* cinemagraph or predicting a ranking for a specific user. To this end, we investigate a computational model to predict human perceptual preference for cinemagraphs. This model is trained on rating scores we collected from a user study.

### 4.1. User Study

Our study consisted of a dataset of 459 cinemagraphs,[6] of which mean video length is about 1 sec. The 459 cinemagraphs are the multiple candidates generated from 244 input video clips. The study consisted of 59 subjects; each was shown one cinemagraph at a time in random order, which is loop play-backed until a user provides a rating from 1 to

---
[6]As we are only interested in understanding semantic and subjective preference, we chose cinemagraphs that did not have any significant visual artifacts, so as not to bias the ratings.

5 using the following guideline: 1) rate each cinemagraph based on interestingness/appeal of the cinemagraph itself, 2) if it is not appealing at all (*i.e.*, you would delete it in an instant), rate it a 1, 3) if it is extremely appealing, (*i.e.*, you would share it in an instant), rate it a 5, 4) otherwise, give intermediate scores according to your preference. Before starting the study, each user was instructed, and carried out a short pilot test. In a pilot study, we found that asking users to rate all cinemagraphs was too fatiguing, which affected the rating quality over time. Instead, in our final user study, we limit the total time spent to 20 mins. On average, each subject ended up rating 289 cinemagraphs.

We conducted a simple statistical analysis to see the characteristics, which suggests that user rating behaviors are very diverse in terms of rate distribution shapes and little consensus among users for each cinemagraph. For instance, 72.66% of cinemagraphs in the dataset have the rates of the standard deviation $\sigma > 1$ among users, while the ones having $\sigma < 0.5$ is actually close to 0%,[7] implying strong personal preference for cinemagraphs. Thus, user-dependent preference may not be modeled using a single model across all users (refer to *global model*). We instead learn a local preference model for each user. In addition, we have to handle partial information, since every subject rated only about 63% of all the cinemagraphs.

### 4.2. Preference Prediction Model

Given the user-study data, our goal is to predict subjective preference rating for a user. A basic approach we can consider is to model subjective preference by associating a regression model to each user independently (refer to *individual model*). However, it is not practical due to two issues on this model: (1) for a new user, we need to train a new model from the scratch, and (2) it requires a lot of data for each user to achieve reasonable generalization. To handle these issues, we use a collaborative style learning to process multi-user information.

To develop a model depending on user and context (cinemagraph), we formulate the problem as a regression, $y = f(\mathbf{v}, \mathbf{u})$, where $y$, $\mathbf{v}$ and $\mathbf{u}$ denote a predicted rating, context and user features respectively. In what follows, we describe the context and user features, and the model $f$.

**Context Feature** The context feature $\mathbf{v}$ can be easily extracted from cinemagraphs, which may be relevant to its preference. We use and concatenate three types of features: hand designed, motion, and semantic features. The hand designed feature consists of quantities related to face, sharpness, trajectory, objectness and loopability. For the motion, we use C3D [36], which is a deep motion feature. For the semantic feature, we use two semantic label occurrence measures for static and dynamic regions. These detail

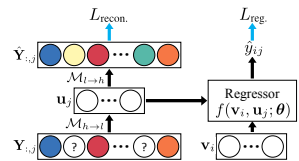

Figure 4: Diagram for architecture and variable dependency of the proposed joint model. The left and right towers denote an auto-encoder and a regression model for rate prediction, respectively.

specifications and lists refer to the supplementary.

**User Feature** Contrary to the context feature, it has not been researched which and what user profiles are related to user's preference for cinemagraph, *i.e.*, undefined. In this regard, we do not use any explicit profile, *e.g.*, age, gender, but instead we leverage rating behavior to reveal user's latent feature. Motivated by collaborative learning [32, 18], we assume that a user's characteristics can be modeled by similar preference characteristics of other users and so it is for similar cinemagraphs. We observed that this is also valid for our scenario (through evidences in Sec. 5 and the supplementary). This allows us to model group behavior and to obtain compact user representation from user rate data without any external information.

We are motivated by an unsupervised approach using auto-encoder [16] to learn the latent user feature such that users with similar preferences have similar features. It has known to have an implicit clustering-effect by enforcing embedding of data to be low-dimensional, called bottleneck [14]. Formally, we represent the multi-user rating information as a matrix $\mathbf{Y} \in \mathbb{R}^{m \times n}$ with $m$ cinemagraphs and $n$ users, of which entry $y_{ij}$ is a rate $\{1, \cdots, 5\}$ of $i$-th cinemagraph by $j$-th user. Given a rating vector for a user, $\mathbf{y}_j = \mathbf{Y}_{:,j}$,[8] we consider two mappings $\{\mathcal{M}\}$ for the auto-encoder, one of which maps a high-dimension vector to low-dimensional space,[9] as $\mathbf{u} = \mathcal{M}_{h \to l}(\mathbf{y})$ and the other is the inverse map as $\mathbf{y} = \mathcal{M}_{l \to h}(\mathbf{u})$. Thus, the auto-encoder can be trained by minimizing self-reconstruction loss, $\|\mathbf{y} - \mathcal{M}_{l \to h}(\mathcal{M}_{h \to l}(\mathbf{y}))\|$. Through this procedure, we can obtain the latent user feature $\mathbf{u}$ from the intermediate embedding. Unfortunately, this is not directly applicable to our problem due to incomplete data (partial ratings by a user). Thus, we leverage a model suggested by Carreira *et al.* [6], *i.e.*, an auto-encoder with missing values (AEm), depicted as the left tower in Fig. 4, whereby rating vectors with missing values are completed and simultaneously non-linear low-dimensional embeddings of rating vectors are learned. Now, we have the latent user feature $\mathbf{u}$. The mappings for $\{\mathcal{M}\}$ make use of a Gaussian radial basis function (RBF) network [4] as suggested by Carreira *et al.*

**Model 1) A Simple User Aware Model** Since we have the described features $\mathbf{u}$ and $\mathbf{v}$, now we can train a regression model such that $y = f(\mathbf{v}, \mathbf{u})$. For the simple baseline model, we use the random forests (RF) regression [11] as a regres-sion function $f(\cdot)$. The RF model is proper for this purpose in that we have limited amount of training data. We use 10 number of ensembles for generalized performance. We call this model as subjective aware RF (S-RF).

---
[7]Statistics of user ratings can be found in the supplementary due to space limitation.

[8]We borrow a MATLAB like matrix-vector representation.

[9]$\mathcal{M}$ applies vector-wise to each column.

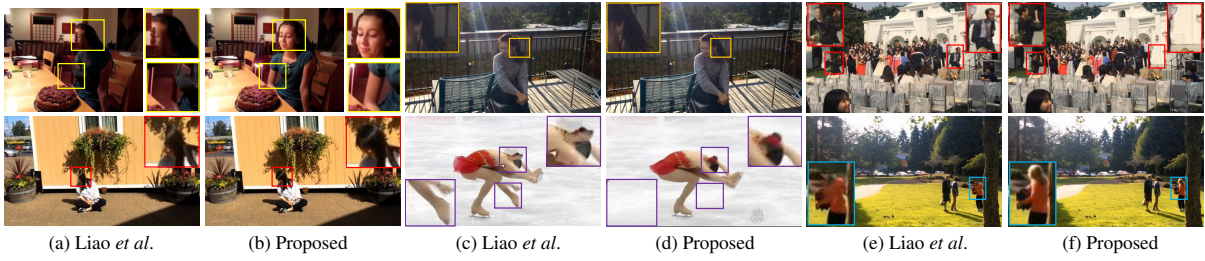

(a) Liao *et al.*   (b) Proposed   (c) Liao *et al.*   (d) Proposed   (e) Liao *et al.*   (f) Proposed

Figure 5: Comparisons with Liao *et al.* [23]. The shown sampled frames are from the cinemagraphs generated by each method. We can observe severe artifacts such as distorted or tearing faces or bodies in (a,c,e), while ours shows artifact-free and semantic preserving results.

**Model 2) A Joint and End-to-End Model**  When we learn **u** by Carreira *et al.* [6], the context feature information is not used; hence, any link between the user and context information may not be reflected to **u**. To learn **u** reflecting context information, we formulate a *joint model* for both regression and auto-encoder that are entangled by user latent feature as a medium variable, of which loss is defined as

$$\arg\min_{\mathbf{U},\mathbf{Y}_{\overline{\Omega}},\{\mathcal{M}\},\theta} L_{\text{reg.}}(\mathbf{U},\boldsymbol{\theta}) + \lambda L_{\text{recon.}}(\mathbf{U},\mathbf{Y}_{\overline{\Omega}},\mathcal{M}_{h\to l},\mathcal{M}_{l\to h}), \quad (8)$$

where $\Omega$ denotes the index set for known entries and $\overline{\Omega}$ is its complementary set, *i.e.*, missing entries, $\mathbf{U} = [\mathbf{u}_1,\cdots,\mathbf{u}_n]$, $\theta$ denotes regression model parameters, and $\lambda = \frac{1}{nm}$ is the balance parameter. $L_{\text{reg.}}$ and $L_{\text{recon.}}$ are respective common $l_2$ regression loss and the the auto-encoder loss of AEm by Carreira *et al*. As with Carreira *et al*., $L_{\text{recon.}}(\cdot)$ incorporates missing values,[10] and defined as:

$$L_{\text{recon.}}(\mathbf{U},\mathbf{Y}_{\overline{\Omega}},\mathcal{M}_{h\to l},\mathcal{M}_{l\to h}) = \quad (9)$$
$$\|\mathbf{U} - \mathcal{M}_{h\to l}(\mathbf{Y})\|_F^2 + \|\mathbf{Y} - \mathcal{M}_{l\to h}(\mathbf{U})\|_F^2 + R_{\mathcal{M}}(\mathcal{M}_{h\to l},\mathcal{M}_{l\to h}),$$

[10]Note that we assume there is no case where all entries in a column vector **y** are missing.

where $\|\cdot\|_F$ denotes Frobenius norm and $R_\mathcal{M}(\cdot,\cdot)$ is the $l_2$ regularization term for two mappings. The same user feature **U** is also fed into $L_{\text{reg.}}(\cdot)$:

$$L_{\text{reg.}}(\mathbf{U},\boldsymbol{\theta}) = \sum_{(i,j)\in\Omega}(y_{ij} - f(\mathbf{v}_i,\mathbf{u}_j,\boldsymbol{\theta}))^2 + R_f(\boldsymbol{\theta}), \quad (10)$$

where $R_f(\cdot)$ is the $l_2$ regularization term for the rating regressor $f$, and we use a linear regression for $f(\cdot)$ as $f(\mathbf{u},\mathbf{v},\boldsymbol{\theta}) = \boldsymbol{\theta}^\top[\mathbf{u};\mathbf{v};1]$. The variable dependency and overall architecture are shown in Fig. 4. We optimize Eq. (8) by the Gauss-Newton method in an alternating strategy. Its optimization details can be found in the supplementary.

Having two loss functions on the same rating may seem redundant, but the information flow during optimization is significant. The sum of two gradients, back-propagated through the rating regressor $f(\cdot)$ to **U** (see Fig. 4) and from $L_{\text{recon.}}$, encourages **U** to be learned from auto-encoding with missing completion and context aware regression. This can be regarded as multi-task learning, which has regularization effect [34] that mitigates the problems of partial and limited number of measurements. This is because it collaboratively uses all the ratings provided by all the users, whereas the *individual model* does not.

For new user scenario, it can be dealt with in a way similar to [18, 37, 31] by finding a similar other user in database.

## 5. Results

**Implementation and Run-time Speed**  We implemented our approach on a PC with 3.4GHz CPU, 32GB RAM and NVIDIA GTX1080 GPU, and applied it over a hundred of casually shot videos acquired by ourselves or collected from previous works. For speed purposes, we downsampled and trimed the videos so that the maximum resolution does not exceed 960×540 and the duration is less than 5-sec long. Without careful engineering level code optimization, our system typcially takes a few minutes to generate all the semantically meaningful cinemagraphs for each video.

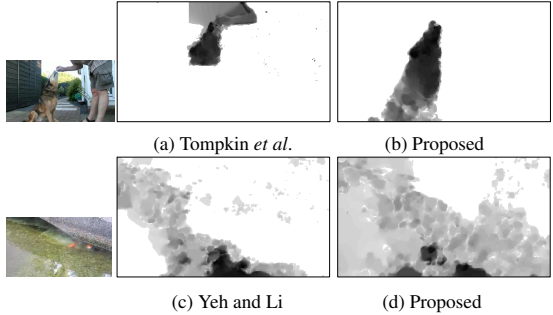

(a) Tompkin *et al.*   (b) Proposed

(c) Yeh and Li   (d) Proposed

Figure 6: Comparison with Tompkin *et al.* [35] and Yeh and Li [39]. The intensity maps indicate average magnitude of optical flow (darker represents larger magnitude). The dynamic areas in our results are better aligned along semantic boundaries of moving objects ("animal" in (a,b), "water" in (c,d)), than other methods.

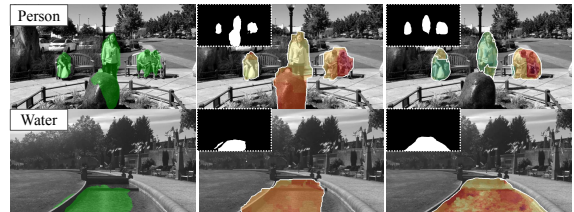

Figure 7: Comparison without/with user editing for our method. [Left] Sampled frames overlaid with semantic segmentation mask for a selected object by green color, [Middle] Color coded $\{\mathbf{s}\}$ label obtained by our method without user editing. [Right] Results with user editing. Each superposed black-white mask shows a semantic binary map $\pi_{(\cdot)}$, on which user edits. Color coding of $\{\mathbf{s}\}$ is referred to Fig. 3.

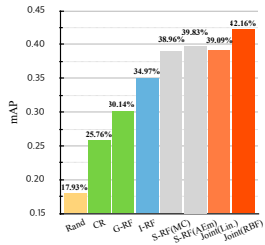

Figure 8: mAP comparison for rating prediction. `Rand`: random guess, `CR`: constant prediction with rate 3, `{G,I,S}-RF`: {global, individual, subjective} RFs, `Joint`: Joint model with either linear or RBF mappings. `MC` and `AEm` indicates user feature obtained from either matrix completion [5] or AEm [6].

**The Importance of Semantic Information** As we argued before, semantic information plays a key role in the process of candiate generation to suppress any semantic meaningless video loops. Thanks to our novel semantic aware cost function (described in Sec. 3) embedded in the MRF framework, the generated cinemagraphs all trend to be more meaningful compared with the ones generated by previous work such as [23] in which only low-level information is considered. Fig. 5 shows a few typical videos that semantic information is crucial to avoid severe artifacts. As indicated by the comparison, the results for Liao *et al*. tend to have artifacts such as distortions or ghosting effects, as highlighted in the close-up views, while our method preserves the boundary region of objects well with more natural looping. Figs. 3 and 6 show another examples of what happens if semantic-based looping is not applied.

**The Effectiveness of Callaborative Learning** We compare the several baselines for cinemagraph preference prediction in Fig. 8 in terms of mean average precision (mAP). Interestingly, `S-RF` and `Joint` outperform `I-RF` (individual learning per a user), which suggests collaboratively learning the preference behavior is beneficial. The best performance of `Joint` shows learning the user feature in a context aware manner can improve the quality of preference prediction for cinemagraph. Another example in Fig. 9 shows the completed rating matrix for missing entries by a matrix completion (MC) [5] (as a reference that does not use context feature) and ours. The completed regions in each left bottom region of matrices clearly show that our method predicts preference ratings more plausibly and diversely than MC by virtue of context aware feature. We visualize 2D embedding of latent user features by t-SNE [26] in Fig. 10, which suggests that users can be modeled by a few types for cinemagraph preference. Refer to supplmentary material for additional results.

**User Interaction** We have showed our results in cases where semantic segmentation worked well. While significant progress has been made on semantic segmentation, the semantic segmentation that we use does not always produce object regions with perfect boundaries or labeling as shown in Fig. 7-[Left], which produces loop labels violating the semantics of the scene (Fig. 7-[Middle]). Using a more advanced semantic segmentation approach such as [13, 12, 9, 8] is one way to improve. However, with simple manual interaction to roughly correct the mask $\pi_\mathbb{D}$, we can quickly fix the issues and output semantically meaningful cinemagraphs (Fig. 7-[Right], where each example took about 19 sec. on average for the editing). This simple optional procedure is seamlessly and efficiently compatible to our MRF optimization (details in the supplementary).

## 6. Discussion and Future Work

We create cinemagraphs using a semantic aware perpixel optimization and human preference prediction. These allow our method to create cinemagraphs without user input; however, the automatic results are limited by the quality of the semantic segmentation. Semantic segmentation itself remains a open research issue beyond the scope of this work, and as these methods improve, they can be used in our approach to improve the results. As an alternative, we optionally allow the user to correct imperfections of semantic segmentation and thus improve the quality of the output. Our system is flexible in that the semantic segmentation part can be seamlessly replaced with an advanced or heterogeneous (*e.g.*, face segmentation) one to improve semantic knowledge or speed, *e.g.*, [28].

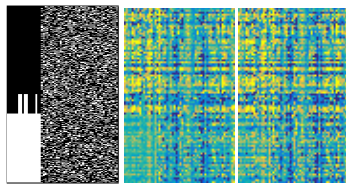

(a) Missing pattern of rates  (b) MC [5]  (c) Joint (RBF)

Figure 9: Completed rating matrices (rows: cinemagraphs, cols.: users). White color indicates missing entries, and rate scores are color-coded through the *parula* color map built in MATLAB.

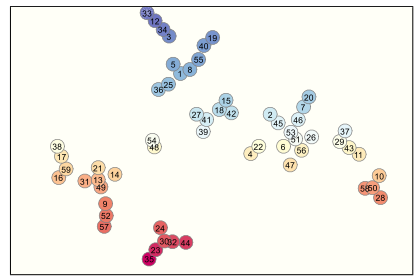

Figure 10: t-SNE visualization for 59 latent user features $\{\mathbf{u}\}$ obtained by `Joint(RBF)`. This plot clearly shows clustered positions of users, which may imply that the intrinsic dimensionality of user space holds the low-dimensionality assumption.

# Supplementary Material:
# Personalized Cinemagraphs using
# Semantic Understanding and Collaborative Learning

Tae-Hyun Oh[1,2]*   Kyungdon Joo[2]*   Neel Joshi[3]   Baoyuan Wang[3]   In So Kweon[2]   Sing Bing Kang[3]
[1]MIT CSAIL, Boston, MA   [2]KAIST, South Korea   [3]Microsoft Research, Redmond, WA

## Summary

This is a part of the supplementary material. The contents of this supplementary material include user study information, implementation details including parameter setups, additional results for the cinemagraph generation and the human preference prediction, and supplementary tables, which have not been shown in the main paper due to the space limit. The supplementary material for resulting videos (comparison with other methods [10, 6, 5, 12], user editing effects, qualitative results) can be found in the project web page.

## 1. User Study Information

During the user study, each cinemagraph is replayed again and again until a user provides a rating for it. The user spends about 4 seconds per cinemagraph on average (we did not limit the time for individual samples but limit the total time by about 20 min.). Before starting the user study, each user was instructed by us, and carried out short pilot tests. The users used the interface provided by us as shown in Fig. 1. On user demographics, the age range is 23-35 years old. About 85% were engineering students and researchers, with the others being non-technical people.

The preference rating could be regarded as an open-ended question. Since the relationships between specific features and user's cinemagraph preference have not been studied, we do not limit any specific preference criteria to avoid bias but capture natural behaviors.

**Statistics of User Ratings**  Fig. 2-(a) shows rating distributions for a random sample of users. The graph shows a very diverse set of rating distributions; the skew and shapes are all quite different. Some of users have a fairly uniform distribution for their ratings, while others clearly favor a certain value (even though few users strongly biased, their ratings are still distributed and express preferences to some extent).

Fig. 2-(b) shows a measure of the diversity of user rating

---
*The first and second authors contributed equally to this work.

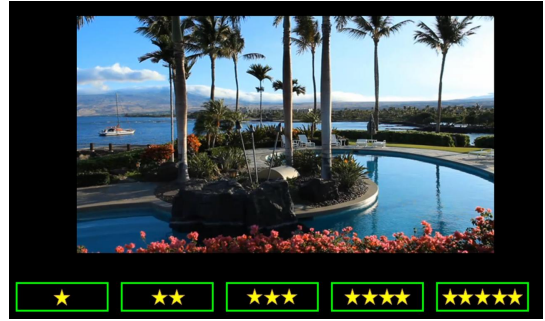

Figure 1: Interface for our user study. The subject is asked to rate a randomly shown cinemagraph (from 1 star to 5 stars).

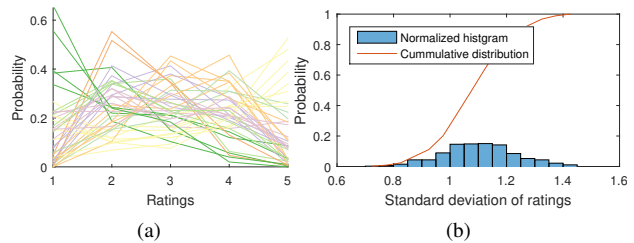

(a)  (b)

Figure 2: Statistics of user ratings. (a) Rating distributions for sampled users (color encoded by clustering users having similar distribution). Different users have diverse tendencies for providing ratings. (b) Distribution of standard deviations of ratings for each candidate cinemagraph across users. Normalized histogram of the standard deviation and its cumulative distribution are overlaid.

scores per cinemagraph. For each candidate cinemagraph, we measure the standard deviation $\sigma$ of user ratings. The histogram in Fig. 2-(b) is constructed by binning the standard deviations for all cinemagraphs.

If the histogram has a pick at $\sigma = 0$, it means all the users gave the same rates for all the cinemagraph, *i.e.*, perfect consensus by common sense. The way of analyzing data may not be same with any traditional statistical test, the presented statistic plot actually implies subjectivenss of rating behavior for cinemagraph. Looking at the overlaid, cumulative distribution curve, it is interesting to see that 72.66% of cinemagraphs in the dataset have $\sigma > 1$, while the percentage of cinemagraphs having $\sigma < 0.5$ is actually close to $0\%$. This represents the diversity of rating tendencies that are user-dependent for a cinemagraph.

## 2. Implementation Details

In this section, we provide the detail information that allows to reproduce our implementation.

**Parameter Setup** All the parameters used in our experiments are listed in the following table:

| Related terms | Parameters |
|---|---|
| $\pi(x)$ in Sec. 3.1 | $thr. = 0.15T$ |
| $E_{\text{label}}$ in Sec. 3.2 | $\alpha_1 = 1$ |
| $E_{\text{spa.}}$ | $\alpha_2 = 15$ |
| $E_{\text{temp.}}$ and $E_{\text{spa.}}$ | $w = 0.2$ |
| $\gamma_t(x)$ in $E_{\text{temp.}}$ | $\lambda_t(x) = \begin{cases} 125, & \text{if } (\vee_{i \in \mathcal{H}_{\text{nat.}}} \pi_i(x)) = 1, \\ 125/2, & \text{otherwise.} \end{cases}$ |
| $\gamma_s(x, z)$ in $E_{\text{spa.}}$ | $\lambda_s = 10/\sqrt{K}$ |
| $E_{\text{label}}$ | $\alpha_\infty = 1000$ |
| $E_{\text{label}}$ | $P_{\text{short}} = 20$ |
| $E_{\text{static}}$ | $\lambda_{\text{sta.}} = 100$ |
| $E_{\text{static}}$ | $\alpha_{\text{sta.}} = 0.03$ |
| $N(\cdot)$ (Gaussian kernel) in $E_{\text{static}}$ | $\sigma_x = 0.9$ and $\sigma_t = 1.2$ |

**Candidate Cinemagraph Generation** The procedure for MRF optimization is as follows:

1. For each looping period label $p>1$, we solve Eq. (1) only for the per-pixel start times $s_{x|p}$ given the fixed $p$, saying $L|p$, by solving a multi-label graph cut with the start frame initialization $s_{x|p}$ that minimizes $E_{\text{temp.}}$ per pixel independently.

2. Given a candidate object label $\mathbb{ID}$, we solve for per-pixel periods $p_x \geq 1$ that define the best video-loop $(p_x, s_{x|p_x})$ where $s_{x|p_x}$ is obtained from the stage (1), again by solving a multi-label graph cut. In this stage, the set of labels are $\{p>1, s'_x\}$, where $s'_x$ denotes all possible frames for the static case, $p=1$.

3. Due to the restriction of the paired label, $s_{x|p}$, in the stage (1), the solution can be restricted. In this stage, we fix $p_x$ from the stage (2) and solve a multi-label graph cut only for $s_x$.

Conceptually, we should alternate the stages (2) and (3). However, in practice, we need to perform the optimization only once, and even then it produces a better solution than the two-stage approach suggested by Liao *et al*. The other difference over Liao *et al*. is that since we generate several candidate cinemagraphs (each representing a different semantic object), we must solve the multi-label graph cut several times.

In MRF optimization, we parallelize the graph cut optimizations using OpenMP and only use a few iterations through all candidate $\alpha$-expansion labels. We find that two iterations are sufficient for the stage (2) and a single iteration is sufficient for all the other stages. To reduce computational cost, we quantize the loop start time and period labels to be multiples of 4 frames. We also set a minimum period length of 32 frames.

**User Editing** To edit the cinemagraph, the user selects a candidate class $\mathbb{ID}$ and a representative frame having regions in which bad boundaries occur.[1] Then, the boundary shape of binary map $\pi_{\mathbb{ID}}$ is edited on overlaid selected frame.

Once the editing is done, the edited $\pi_{\mathbb{ID}}$ is fed into MRF optimization and re-run the stages (2, 3) in the Algorithm 1 with the parameter $\alpha_{\text{sta.}}$ in $E_{\text{static}}(\cdot)$ being doubled, so that the edited regions are strongly encouraged to be dynamic. Note that despite increasing $\alpha_{\text{sta.}}$, a non-loopable region will remain static. Rerunning the stages (2, 3) requires initialization and pre-computed $\{s_{x|p}\}$, but we can re-use these pre-computed quantities from the stage (1).

**Context Feature** For the context feature, we use three types of features: hand designed, motion, and semantic features. We extract 55-dimensional hand designed features, which consist of face, sharpness, trajectory, objectness and loopability (its details are listed in Sec. 4 of this supplementary material). We use C3D [11] as the motion feature, which is a deep motion feature obtained from 3D convolutional neural network. We apply C3D with the stride of 16 frames and 8 frame overlap, and average pooling is applied, so that we have a 4096 dimensional representative motion feature for each cinemagraph. For the semantic feature, we use two semantic label occurrence measures for static and dynamic regions as $\vec{h}_{\text{static}} = \sum_{x \in \text{static}} \vec{h}(x)$ and $\vec{h}_{\text{dyn.}} = \sum_{x \in \text{dynamic}} \vec{h}(x)$ respectively. The final context feature for a cinemagraph is formed by concatenating all the mentioned feature vectors, where each feature is independently normalized by the infinity norm, *i.e.*, the largest absolute value, before concatenation.

**Model 2) A Joint and End-to-End Model** We apply an alternating optimization strategy iteratively over $(\mathbf{U}, \mathbf{Y}_{\overline{\Omega}})$ and $(\{\mathcal{M}\}, \boldsymbol{\theta})$; we first fix $(\{\mathcal{M}\}, \boldsymbol{\theta})$ during optimizing $(\mathbf{U}, \mathbf{Y}_{\overline{\Omega}})$ and followed by $(\{\mathcal{M}\}, \boldsymbol{\theta})$ while fixing $(\mathbf{U}, \mathbf{Y}_{\overline{\Omega}})$ until convergence. When fixing $(\{\mathcal{M}\}, \boldsymbol{\theta})$, optimizing $(\mathbf{U}, \mathbf{Y}_{\overline{\Omega}})$ is the non-linear least square problem. We optimize it using the Gauss-Newton method, where $\frac{\partial f(\mathbf{u}, \mathbf{v}; \boldsymbol{\theta})}{\partial \mathbf{u}}$ is added when updating $\mathbf{U}$. In the process of minimizing $L_{\text{recon.}}$, missing values $\mathbf{Y}_{\overline{\Omega}}$ are regarded as optimization variables while $\mathbf{Y}_{\Omega}$ is kept constant.

When we solve for $(\{\mathcal{M}\}, \boldsymbol{\theta})$, we separately solve three regressions for $\{\mathcal{M}\}$ and $f(\cdot; \boldsymbol{\theta})$. The mappings for $\{\mathcal{M}\}$ use Gaussian radial basis function (RBF) network [2] to provide a non-linear mapping, $\mathcal{M}(\mathbf{x}) = \mathbf{W}\mathcal{K}(\mathbf{x})$ where $\mathcal{K}(\mathbf{x}) = [\kappa_1(\mathbf{x}), \cdots, \kappa_d(\mathbf{x})]$ ($d \ll \min(m, n)$), where $\mathbf{Y} \in \mathbb{R}^{m \times n}$, and $\kappa_i(\mathbf{x}) = \exp(\frac{1}{2\sigma_i^2} \|\mathbf{x} - \mu_i\|_F^2)$.[2] For the regressions for $\{\mathcal{M}\}$ between $\mathbf{U}$ and $\mathbf{Y}$, we update respective $\{\mu\}$ by $k$-means and $\{\sigma\}$ by cross validation with a subset that is split from the training set used for RBF training. Then, $\{\mathbf{W}\}$ is solved for by a least square fit. With this RBF mapping, the regularization term is defined as

$$R_{\mathcal{M}}(\mathcal{M}_{h\to l}, \mathcal{M}_{l\to h}) = \lambda_R \left( \|\mathbf{W}_{h\to l}\|_F^2 + \|\mathbf{W}_{l\to h}\|_F^2 \right).$$

---

[1] Since it is used as a guide, it does not have to be exact.
[2] When we use a linear mapping for $\mathcal{M}$, it reduces to a linear model that forms matrix factorization.

The rating regressor $f(\cdot)$ uses a linear function as $y = f(\mathbf{u}, \mathbf{v}; \boldsymbol{\theta}) = \boldsymbol{\theta}^\top [\mathbf{u}; \mathbf{v}]$. Again, the parameter $\boldsymbol{\theta}$ is updated by least square fit with its regularization term $R_f(\boldsymbol{\theta}) = \lambda_\theta \|\boldsymbol{\theta}\|_F^2$. The regularization parameters are set as $\lambda_R = \lambda_\theta = 0.1$. The number of RBF basis functions is set as $d = 25$. These parameters are chosen by running the algorithm on the separated validation set (more details are described in Sec. 3.2 of this supplementary material), which was not used for test in all experiments. We use a validation dataset for parameter tuning with the parameter sets $\lambda_R = \lambda_\theta = \{1e^{-6}, 1e^{-5}, 1e^{-4}, 1e^{-3}, 1e^{-2}, 0.1, 1\}$ and $d = \{5, 10, 15, 20, 25, 30, 35, 40, 45\}$.

In our method, we initialize $\mathbf{Y}_{\overline{\Omega}}$ from the convex matrix completion (MC) [3] with speeding up by [9], $\mathbf{U}$ from Laplacian eigenmap [1] on $\mathbf{Y}$ obtained from MC with 25 dim as mentioned above. Then, with this initialization, the mappings $\{\mathcal{M}\}$ and the rating regressor $f(\cdot)$ are fit.

## 3. Additional Results

In this section, we present additional qualitative results for semantic cinemagraph generation, followed by extensive evaluation on the computational model for human preference prediction.

### 3.1. Evaluation on Semantic Cinemagraph Generation

**Computational Time Profile** In our experiments, the input videos are at most 5 seconds long, with maximum rate of 30 frames/second. The resolution is at most $960 \times 540$ pixels; higher resolutions are down-sampled. The processing time for a 3-sec $960 \times 540$ video takes a few minutes, depending on the number of candidates. Here is the breakdown in timing: initialization $\approx$10 secs (the stage (1) in Algorithm 1), MRF solving $\approx$50 secs per candidate (the stages (2, 3) in Algorithm 1), and rendering $\approx$10 secs.

**Additional Qualitative Comparison** Figure 3 shows a comparison with Tomkin *et al*. [10]. The method of Tomkin *et al*. allows user to select the region and loop to be animated, but has no synchronization feature. The example of Tomkin *et al*. have not only the desynchronized animation on eye blink and visual artifacts on that region, which shows what happens if semantic-based looping is not applied. The differences are clearer in our supplementary video, which we encourage the reader to view.

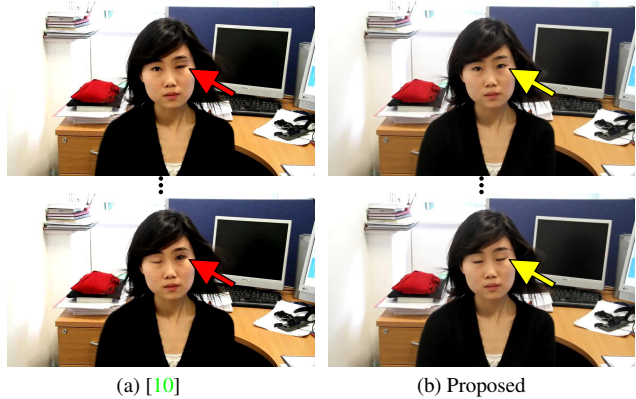

(a) [10]      (b) Proposed

Figure 3: Comparisons of our cinemagraph generation with Tompkin *et al*. [10]. In the result (a) of Tompkin *et al*., although a winking effect on the eyes is intentionally introduced by user editing, it generates unsynchronized one (red arrow) with visual artifact, while our result in (b) shows synchronized eye blinking of person (yellow arrow).

**Cinemagraph Visualization** Figure 4 shows representative examples of cinemagraphs rendered using different periods and start frames ($\{\mathbf{p}, \mathbf{s}\}$ respectively). Each row is of the same scene, and each column represents a candidate cinemagraph (*i.e.*, a different object to animate). The heat map indicates how dynamic the region is, with gray being static. The preference prediction results in Fig. 4 will be explained in the subsequent section.

### 3.2. Evaluation on Human Preference Prediction

In this section, we evaluate the preference prediction model described in Sec. 4 of the main paper in the following ways: performance and visualization of grouping effect. Throughout our experiments, we randomly sampled 10% rating data as the validation set, and tune parameters of methods using this set. We use the rest of the data for 9-fold cross validation, so that the amount of test set is same with the validation set.

**Performance** In Fig. 8 of the main paper, we consider other regression methods to understand the effects of several factors, and especially choose randomized forests (RF) [4] as the main competitor.[3] Fig. 8 of the main paper shows the performance comparison: `Rand`: random guess (a lower bound of the performance), `CR`: constant prediction model with rate 3, `G-RF`: a single global RF model for all users, `I-RF`: RFs individually learned for each user, `S-RF+{MC, Ours}`: a single RF model for all users with subjective user feature obtained from either `MC` or `Ours` (for both user features, we use 25 dimensions), `Ours`: the proposed method with either linear or RBF mapping func-tions. `G-RF` and `I-RF` require context feature only, while `S-RF`s require both context and user features. For RF based methods, we use 10 number of ensembles.

---

[3] We tested other regression methods, such as linear, support vector, Gaussian process, multi-layer perceptron, for the rate prediction given context and user features. In our scenario with limited amount of training data, RF performed best; hence we only report RF based results for simplicity.

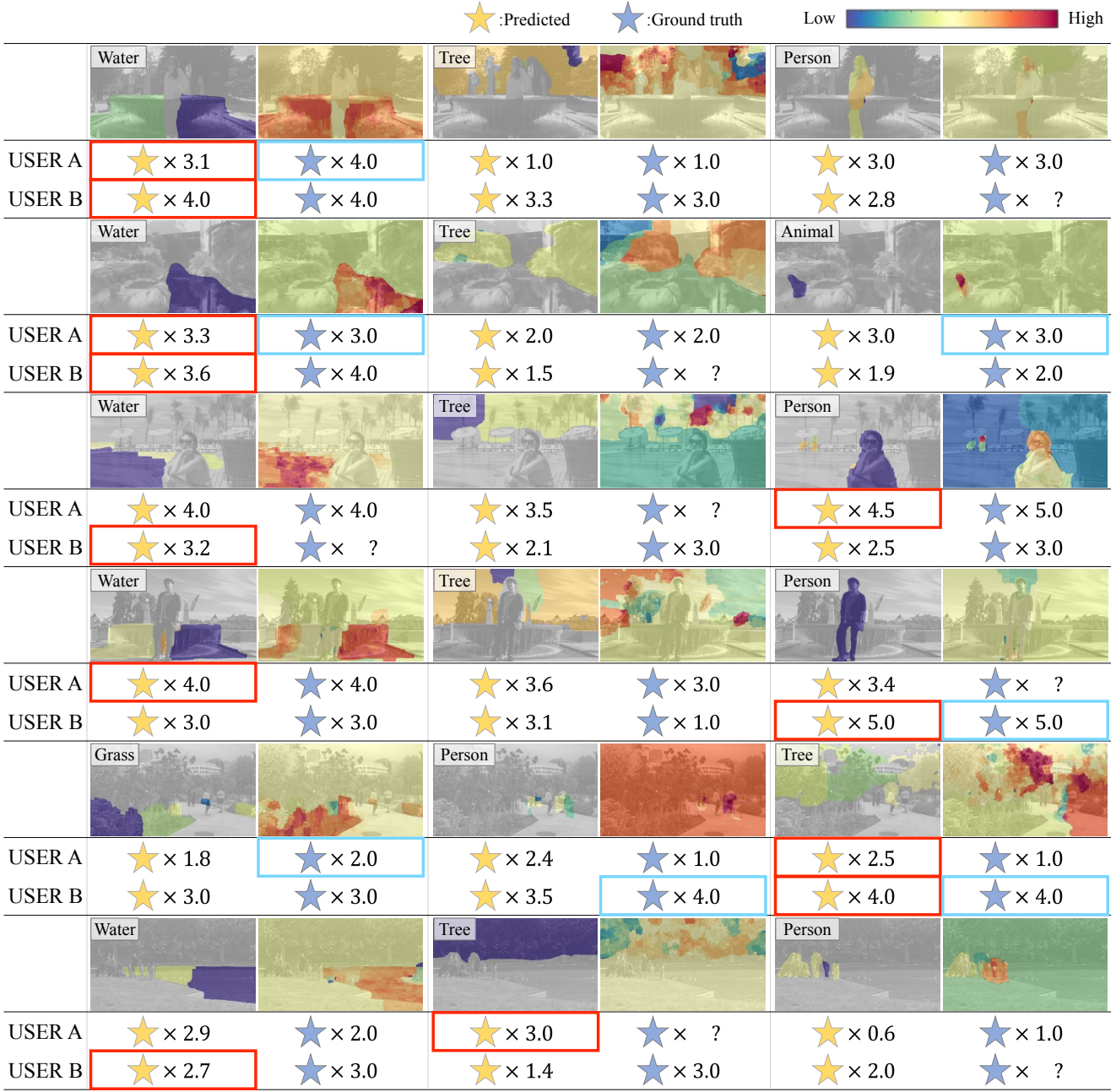

Figure 4: Visualization of $\{\mathbf{p}, \mathbf{s}\}$ and predicted ratings for unseen cinemagraphs by our prediction model. Each row presents three different candidate cinemagraphs generated from a single video input, and subsequent two columns are a pair of $\{\mathbf{p}\,(\text{left}), \mathbf{s}\,(\text{right})\}$, whose value is presented by a color map ranging from blue to through yellow to red as values increase, with gray indicating static pixels. Note that the presented cinemagraphs are unseen data during training. Preferences are not observed for every combination of users and cinemagraphs, which is indicated by the symbol '?' as unknown ground truth. Red highlights indicate the selected best cinemagraph for each user according to the predicted preference rates, and blue highlights indicate the true preference according to the surveyed preference rate.

It is worthwhile to see the learnability of human preference by comparing simple regression, *i.e.*, `G-RF` and `I-RF`. As mentioned in Sec. 7.1 of the main paper, we cannot find any common sense from the statistics of user ratings, rather it reveals the fact that users' preferences are too subjective;

it can be deduced from low mAP of `G-RF`. Note that modeling of `G-RF` can be regarded as an attempt to learn a common sense of human preference. In order to show the importance of the user feature, we compare `S-RF`, which uses both user and context features, with `G-RF` and `I-RF`. The improvement of `S-RF` over `G-RF` and `I-RF` clearly shows the importance of the user feature. On the other hand, the importance of context feature is shown by comparing `S-RF` and `MC` which do not use context feature. Notice that `S-RF` can be used only when user feature is given by other methods that can learn user feature in an unsupervised manner such as `MC` or `Ours`. Thus, `S-RF` is an ideal comparison in the setup without given user feature. Nonetheless, `Ours (RBF)` achieves the best performance over `S-RF` by virtue of joint approach to learn user representation and regression. Lastly, comparing to `Ours (Lin.)` shows that the non-linear dimension reduction is crucial for implicit user relational modeling in a collaborative learning regime. Running time of `Ours (RBF)` takes about 72 seconds in unoptimized MATLAB implementation with a matrix of $459 \times 59$.

**Qualitative Examples of the Predicted Rating** We present rate prediction examples in Fig. 4, and highlight the selected best cinemagraph for each user by colors. Note that the presented cinemagraphs are unseen data during training. Since preferences are not observed (surveyed) for every combination of users and cinemagraphs, unknown ground truth is indicated by the symbol '?'. It is well reflected by the proposed method that each user has their own subjective for best preferred cinemagraph, and overall the predictions have good matches with the selected best cinemagraphs by ground-truth.

**Grouping Effect** Given the user representation obtained by `Ours (RBF)`, we visualize its 2-dimension embedding by t-SNE [7] in Fig. 5. The plot clearly shows clustered positions of users, which may imply that the intrinsic dimensionality of user space holds the low-dimensionality assumption. To see tendencies among neighbor users in the embedding space, we display true ratings of sampled users in Fig. 6. The users and groups are sampled by considering the proximity in the 2D embedding, and the cinemagraphs are sampled from a set in which entries are rated by all the presented users directly (none of them are inferred). The user IDs correspond to the node IDs in Fig. 5. It shows that each group has similar preference tendency, which implies that the users located at similar embedding space have similar preference characteristics.

## 4. Supplementary Tables

**Hand-Designed Feature List** Figure 7 is the hand-designed feature list used in the human preference learning part. The low-level hand designed feature has total 55-

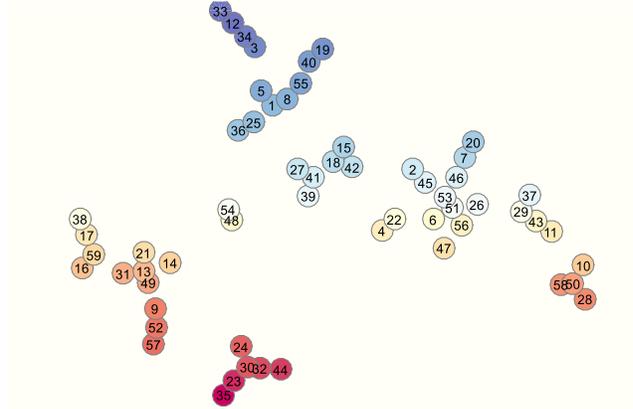

Figure 5: t-SNE visualization for 59 latent user features.

| UIDs<br>Cinema. | Group A | | | Group B | | | Group C | | |
|---|---|---|---|---|---|---|---|---|---|
| | 23 | 30 | 32 | 1 | 5 | 8 | 26 | 51 | 53 |
| | 1 | 1 | 1 | 3 | 1 | 4 | 4 | 4 | 5 |
| | 2 | 1 | 1 | 5 | 5 | 5 | 2 | 3 | 3 |
| | 2 | 2 | 3 | 5 | 5 | 5 | 2 | 3 | 2 |
| | 2 | 3 | 2 | 4 | 1 | 3 | 3 | 3 | 3 |
| | 3 | 1 | 1 | 5 | 5 | 5 | 2 | 5 | 5 |
| | 2 | 1 | 2 | 1 | 1 | 1 | 2 | 1 | 1 |
| | 2 | 3 | 5 | 2 | 1 | 1 | 2 | 2 | 2 |
| | 3 | 4 | 4 | 5 | 5 | 4 | 1 | 3 | 5 |
| | 3 | 4 | 4 | 2 | 1 | 2 | 3 | 2 | 2 |

Figure 6: Group behavior of user preference among intra- and inter-groups. The presented ratings are the numbers directly provided by each user. The users are sampled according to the proximity of embeddings in Fig. 5, and the presented cinemagraphs are sampled as those are rated by all the listed users, *i.e.*, intersection set. Green color overlay indicates dynamic looping regions, otherwise static.

dimension. The presented order of this list is identical to the order of feature vector entries.

**Semantic Class Mapping Table** These semantic classes are based on PASCAL-Context [8]. This class mapping table in Fig. 8 is used to combine some categories and classify natural/non-natural categories in the semantic-based cinemagraph generation method. The dot . in the mapping

| Type | Dimension | Feature |
|---|---|---|
| Face | 15 | facesizeMin<br>facesizeMax<br>facesizeMean<br>facesizeMedian<br>facesizeStd<br>facexsMin<br>facexsMax<br>facexsMean<br>facexsMedian<br>facexsStd<br>faceysMin<br>faceysMax<br>faceysMean<br>faceysMedian<br>faceysStd |
| Texture | 5 | sharpnessMin<br>sharpnessMax<br>sharpnessMean<br>sharpnessMedian<br>sharpnessStd |
| Motion flow | 10 | motionMin<br>motionMax<br>motionMean<br>motionMedian<br>motionStd<br>motionSurroundMin<br>motionSurroundMax<br>motionSurroundMean<br>motionSurroundMedian<br>motionSurroundStd |
| Trajectory | 15 | tracklengthMin<br>tracklengthMax<br>tracklengthMean<br>tracklengthMedian<br>tracklengthStd<br>trackBoundingBoxMin<br>trackBoundingBoxMax<br>trackBoundingBoxMean<br>trackBoundingBoxMedian<br>trackBoundingBoxStd<br>trackTravelsMin<br>trackTravelsMax<br>trackTravelsMean<br>trackTravelsMedian<br>trackTravelsStd |
| Global loopability | 5 | globalLoopCostsMin<br>globalLoopCostsMax<br>globalLoopCostsMean<br>globalLoopCostsMedian<br>globalLoopCostsStd |
| Face ratio | 5 | faceRatiosMin<br>faceRatiosMax<br>faceRatiosMean<br>faceRatiosMedian<br>faceRatiosStd |

Figure 7: Hand-designed feature list used in the human preference learning part. It has total 55 dimension.

class denotes that original class name is used and left intact.

| ID | PASCAL-Context | Mapping class | Natural category |
|---|---|---|---|
| 1 | background | background | natural |
| 2 | aeroplane | . | |
| 3 | bicycle | bike | |
| 4 | bird | animal | |
| 5 | boat | . | |
| 6 | bottle | household item | |
| 7 | bus | . | |
| 8 | car | . | |
| 9 | cat | animal | |
| 10 | chair | chair | |
| 11 | cow | animal | |
| 12 | diningtable | household item | |
| 13 | dog | animal | |
| 14 | horse | animal | |
| 15 | motorbike | bike | |
| 16 | person | person | |
| 17 | pottedplant | grass | natural |
| 18 | sheep | animal | |
| 19 | sofa | chair | |
| 20 | train | . | |
| 21 | tvmonitor | . | |
| 22 | bag | . | |
| 23 | bed | . | |
| 24 | bench | chair | |
| 25 | book | . | |
| 26 | building | background | natural |
| 27 | cabinet | household item | |
| 28 | ceiling | background | natural |
| 29 | clothes | person | |
| 30 | computer | . | |
| 31 | cup | household item | |
| 32 | door | . | |
| 33 | fence | . | natural |
| 34 | floor | background | natural |
| 35 | flower | grass | natural |
| 36 | food | household item | |
| 37 | grass | grass | natural |
| 38 | ground | background | natural |
| 39 | keyboard | . | |
| 40 | light | . | natural |
| 41 | mountain | . | natural |
| 42 | mouse | . | |
| 43 | curtain | . | natural |
| 44 | platform | background | natural |
| 45 | sign | . | |
| 46 | plate | household item | |
| 47 | road | background | natural |
| 48 | rock | . | natural |
| 49 | shelves | household item | |
| 50 | sidewalk | background | natural |
| 51 | sky | sky & tree | natural |
| 52 | snow | water | natural |
| 53 | bedcloth | . | |
| 54 | track | background | natural |
| 55 | tree | sky & tree | natural |
| 56 | truck | . | |
| 57 | wall | background | natural |
| 58 | water | water | natural |
| 59 | window | . | |
| 60 | wood | . | natural |

Figure 8: Class mapping table used in the semantic-based cinemagraph generation method.